\documentclass[runningheads,a4paper]{llncs}

\usepackage{amsmath, amssymb}
\usepackage{graphicx,enumitem, subfigure,url}
\usepackage[boxed,linesnumbered]{algorithm2e}

\begin{document}

\mainmatter  

\title{Towards Single-phase Single-stage Detection of \\ Pulmonary Nodules in Chest CT Imaging}

\titlerunning{Towards Single-phase Single-stage Detection of Pulmonary Nodules}

%
%
 \author{Zhongliu Xie}
 %
  \authorrunning{Zhongliu Xie}
%
 \institute{Imperial College London, \\ London, United Kingdom}
%
%

\maketitle

\vspace{-4mm}

\begin{abstract}
Detection of pulmonary nodules in chest CT imaging plays a crucial role in early diagnosis of lung cancer. 
Manual examination is highly time-consuming and prone to errors, calling for computer-aided detection, both to improve detection efficiency and reduce misdiagnosis. 
Over the years, a large number of such systems have been proposed, which mostly followed a two-phase paradigm with: 1) candidate detection and 2) false positive reduction. Recently, deep learning has become a dominant force in algorithm development. 
As for candidate detection, prior state-of-the-art was mainly based on the two-stage Faster R-CNN framework, which starts with an initial sub-net to generate a set of class-agnostic region proposals, followed by a second sub-net to perform classification and bounding-box regression. 
In contrast, we abandon the conventional two-phase paradigm and two-stage framework altogether, and propose to train a single network to achieve end-to-end nodule detection instead, without transfer learning or further post-processing. 
Our feature learning model is a modification of the ResNet and feature pyramid network combined, powered by RReLU activation.
The major challenge is the condition of extreme inter-class and intra-class sample imbalance, 
where positive samples are overwhelmed by a vast negative pool largely composed of easily discriminative samples and a handful of hard samples.
Direct training on all samples can seriously undermine training efficacy.
We propose a patch-based sampling strategy over a set of regularly updating anchors, which is able to narrow sampling scope to all positives and only hard negatives, effectively addressing this issue. 
As a result, our approach substantially outperforms prior art in terms of both accuracy and speed. 
Finally, the prevailing evaluation method is a FROC analysis over [1/8, 1/4, 1/2, 1, 2, 4, 8] false positives per scan, which is far from ideal in real clinical environments. Regarding practical considerations, we suggest FROC over [1, 2, 4] false positives as a better metric. 



\keywords{Lung Nodule Detection, Deep Learning, Single-phase Single-stage Detection, Anchor-based Patch Sampling}
\end{abstract}

\section{Introduction}
\vspace{-2mm}

Lung cancer is the leading cause of death in both men and women, accounting for around 
1.6 million or 20\% of deaths worldwide every year \cite{2}. 
According to relevant studies \cite{1}, the 5-year survival rate for lung cancer is around 18\%, and over half the cases are diagnosed at a late stage, lowering the rate to only 4\%. 
The situation could be significantly improved by an earlier diagnosis, for instance, through pre-screening of pulmonary nodules with low-dose computed tomography (CT).
Pulmonary nodules are considered a crucial indicator of primary lung cancer.
Under CT imaging, they often manifest as small opaque structures, brighter than surroundings and roughly of spherical shape. 
Once identified, radiologists can then conduct malignancy study, paving the way for early treatment. 
However, a chest CT scan typically generates hundreds of slices. 
Manual examination is highly time-consuming and prone to errors, 
calling for computer-aided detection, both to improve detection efficiency and reduce misdiagnosis. 



Over the years, a large number of such systems have been proposed, which mostly followed a two-phase paradigm with: 
1) candidate detection, and 2) false positive (FP) reduction. 
It starts by a relatively coarse scan over the entire CT volume to locate a considerable set of potential candidates,
typically with high sensitivity but low precision. This is followed by a more rigorous process to screen out FP detections to improve precision.
Traditional candidate detection methods include intensity thresholding, shape curvedness and mathematical morphology \cite{5,7};
whereas traditional FP reduction methods include combinatorial analytics over shape, position, intensity, gradient features, 
texture characteristics and contextual information \cite{5,9}. 
However, all these have only achieved limited success. 

Recent advances have primarily been driven by deep learning with convolutional neural networks (CNN). 
Most notably, Ding et al. \cite{ding} proposed using 2D Faster R-CNN \cite{faster-rcnn} with an ImageNet pre-trained VGG-16 model \cite{VGG} to carry out candidate detection, followed by a 9-layer 3D CNN to perform FP reduction. 
Their methodology scored 0.893 on the well-known LUNA benchmark \cite{luna}, measured by average sensitivity over 7 conditions, respectively [1/8, 1/4, 1/2, 1, 2, 4, 8] FP/scan.
Regarding FP reduction alone, Setio et al. \cite{setio} proposed to incorporate wider spatial information for nodule classification,
by using multi-view CNNs to process 2D patches extracted from different image planes. They achieved 0.854 and 0.901 sensitivities at 1 and 4 FP/scan respectively. 
Moreover, Dou et al. \cite{douqi} suggested using a multi-level contextual 3D CNN to perform classification over three differently sized 3D patches centring each nodule candidate, followed by weighted label fusion. They scored 0.827 in FP reduction on LUNA.

Nevertheless, the prevailing two-phase paradigm requires designing separate methods over a pipeline to combine detection efficacy. 
In the case of deep learning, it generally entails training two completely independent networks, 
which not only introduces considerable engineering efforts, but also neglects that the two phases are highly correlated in nature. 
A more sensible way may be to conduct joint-training over the two networks. 
Moreover, in terms of candidate detection, in Ding et al. \cite{ding} for example, 
the borrowed Faster R-CNN \cite{faster-rcnn} is a popular two-stage detection framework for object recognition.
It starts with an initial sub-net (a.k.a. region proposal network) to generate a set of class-agnostic region proposals, 
followed by a second sub-net to perform classification and bounding-box regression.
However, this methodological design, with respect to nodule detection,  is to some extent, redundant, since both sub-nets produce a similar output: a binary classification and a bounding box regression.




In stark contrast, we abandon the conventional two-phase paradigm and two-stage detection framework altogether, 
and propose to train a single network to achieve end-to-end detection instead, without transfer learning or further post-processing. 
It therefore is very simple and concise: the model extracts features directly from 3D space, and estimates nodule probability and location straight-away. 
As for architectural design, we borrow skip-connection from the residual network (ResNet) \cite{resnet} to address  ``network degradation'' and ``vanishing gradients'' in deep models, which is widely used in both recognition and generation tasks \cite{chen2017stylebank,he2018deep}. To enhance multi-scale detection performance, we adopt pyramidal feature projection with lateral connections from the feature pyramid network (FPN) \cite{fpn}. The network is powered by RReLU \cite{rrelu} activation to combat the problems of ``dying neurons'' and non-zero centred response distribution.



Secondly, the major challenge to train deep networks to accurately perform nodule detection is the condition of extreme sample imbalance (a common issue in many medical imaging problems), which may be further divided to: 
1) inter-class imbalance, where the number of negative samples (non-nodule) far exceeds the positive (nodule); 
2) intra-class imbalance, where most negatives (mainly the air) are easily discriminable whereas the rest few (e.g. blood vessels) is far more challenging. 
Direct training on all samples is inefficient, as the overwhelming presence of easy negatives would dilute training efficacy. 
To address the issue, we propose a simple and efficient training strategy with patch-based sampling over a set of regularly updating anchors (called ``sampling anchors'' to differentiate from the ``regression anchors'' to initialise box regression in Faster R-CNN \cite{faster-rcnn}). 
This strategy significantly narrows sampling scope, reducing to all positives and only hard negatives, effectively combating sample imbalance. 
Moreover, by training over patches that are much smaller than full-sized images, it substantially lowers memory consumption: a concrete constraint with 3D CNNs on today's GPUs.

Finally, the prevailing evaluation method is a FROC analysis that calculates average sensitivity over 7 FP/scan conditions described earlier, which is promoted by the LUNA benchmark and we denote it FROC[luna]. 
Our third contribution marks a pilot exploration towards clinical practice in an on-going follow-up study, in which we observed that FROC[luna] is far from ideal. In real clinical environments, the algorithm is best used as a tool to assist radiologists, who review and revise its output before making a final diagnosis. 
In this case, false negative detections, which are difficult to revise, are much more concerned than false positives, which are much easier. 
Radiologists are generally more tolerant to several FPs in a scan, with positive opinions towards a small trade-off of precision for improved sensitivity. We therefore suggest the FROC over [1, 2, 4] FP/scan as a better evaluation metric, denoted as FROC[new] for comparison.

\section{Methods}


\subsection{Data Pre-processing}
Depending on the imaging procedure,  different CT scans may feature different sizes, spacings and orientations. 
As a preliminary step, all images are re-sampled to  $1 \times 1 \times 1mm$ and adjusted to the same orientation.
Then, pulmonary nodules are distributed in the lungs, which are situated within the thoracic cavity of the human body.
Instead of scrutinising the entire chest scan, cropping it to the lung region can effectively narrow search space.
The lung region mainly contains air, and its volume may expand to over 10 litres during breath-in and contract to 1 litre during breath-out. 
In this case, we apply a simple thresholding regarding the Hounsfield unit of the air (-1000 HU), followed by morphological methods, 
including 3D connected component analysis and region erosion/dilation, to obtain a coarse lung segmentation. 
After that, the image is cropped by the bounding box of lung segmentation, with intensities re-scaled to [0,1] in real numbers. 


\begin{figure}[!b]
\centering
\subfigure{\includegraphics[height = 4.2cm]{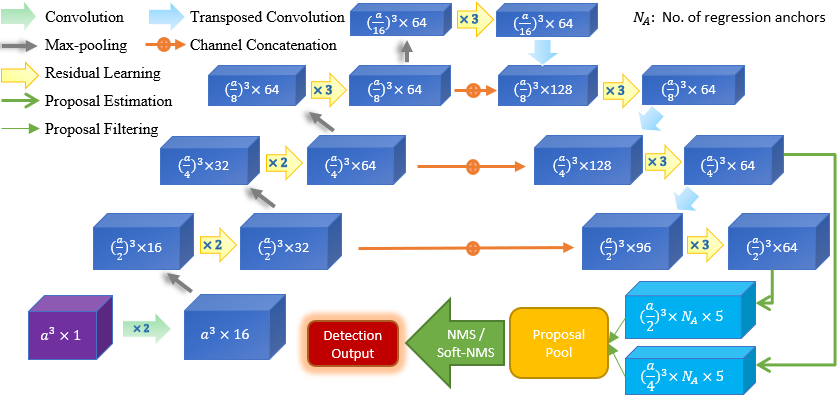}}
\subfigure{\includegraphics[height = 4.2cm]{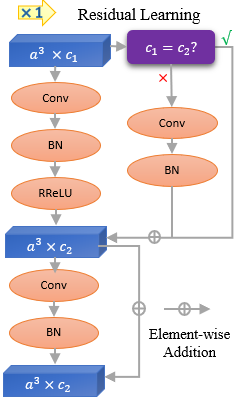}}
\caption{Architecture of the Nodule Detection Framework}
\label{fig:architecture}
\vspace{-5mm}
\end{figure}

\subsection{Nodule Detection Framework}
As we employ a single-phase single-stage approach, only one network is used to perform end-to-end nodule detection.
Its architecture is illustrated in Fig. \ref{fig:architecture}. 
Our feature learning draws insights from the ResNet \cite{resnet} and FPN \cite{fpn}. 
The ResNet utilises skip-connection to enhance information flow over feature propagation through deep networks, able to mitigate the ``network degradation'' and ``vanishing gradients'' problems.
In our case, residual learning is extended to 3D and adopted in the base feature extraction block shown in Fig. \ref{fig:architecture} (right). 

The FPN performs multi-scale feature projection with a pyramidal hierarchy, which consists of a down-sampling and an up-sampling pathway. A similar structure was proposed earlier in the U-Net work \cite{u-net} that won several ISBI challenges a few years back.
Yet in the U-Net, prediction is made on the last feature map only, 
whereas the FPN utilises the semantic feature maps at all scales by lateral connections to a set of output heads making predictions independently.
In fact, the FPN was originally proposed to be incorporated in Stage One of Faster R-CNN \cite{faster-rcnn} to generate class-agnostic region proposals, in order to enhance object classification and localisation in Stage Two.
In our work, it is modified as an end-to-end nodule classifier and localiser, which takes in a 3D image and outputs a set of tuples $\left< n,x,y,z,d \right>$ each indicates a nodule presence, where $n$, $\left<x,y,z\right>$ and $d$ respectively  denote its ``noduleness'',  location coordinate and diameter. Moreover, we reduce lateral connections to only the last two scales of the pyramid, 
as nodules ($\leqslant 30mm$) are considered too small when down-sampled by $\times 8$ or over. 
Proposals estimated by the two output heads are filtered by binary thresholding on ``noduleness'', then aggregated using non-maximum suppression (NMS). In the case of dense nodule distribution, soft-NMS \cite{soft-NMS} may be employed instead.

Furthermore, in many deep learning studies, ReLU activation is often applied to address the well-known ``saturating problem''.
Yet its non-negative activation could lead to ``dying neurons'' and non-zero centred response distribution. 
As a remedy, a number of modifications have been proposed to introduce a non-zero slope in the negative half of axis, for instance,  with a small fixed value (Leaky ReLU), a randomised value (Randomised ReLU, RReLU), or with a tunable parameter (Parametric ReLU). 
The RReLU has been reported with superior performance in several comparative studies \cite{rrelu} and is thus adopted in our study.
\subsection{Network Training with Anchor-based Patch Sampling}

As described earlier, the major challenge to train deep networks to accurately perform nodule detection is the condition of extreme inter-class and intra-class sample imbalance, where the positive samples are sparsely distributed and overwhelmed by a vast negative pool largely composed of easy samples and a handful of hard samples. Direct training on all samples is inefficient, as the prevalence of easy negatives would dilute training efficacy, leading to lower detection accuracy. To address this issue, we propose a simple, effective and efficient training strategy using patch-based sampling over a set of regularly updating anchors. 

In each sampling step, a patch of size $\gamma^{3}$ (e.g. $96^{3}$) is cropped around each anchor and passed over for network inference, which generates a proposal grid (e.g. $24^{3} \times 10 \times 5$). Training samples are then retrieved from the grid, where a sample is considered positive if it overlaps the ground truth with an intersection-over-union (IOU) above $\eta_{+}$, or negative if below $\eta_{-}$. Subsequently, online hard example mining \cite{ohem} is employed, in which only $k$ negatives with highest  ``noduleness'' are passed over for loss calculation and model update. 
The sampling anchors are updated regularly after each training round, which takes a custom $t$ epochs. 
The initial anchors are simply set to the locations of ground truth nodules, i.e. $anchor_{0} = S_{gt}$. Then after each round, a provisional testing is applied to all full-sized images in the training set, where a set of FP detections will be generated and included in anchor settings for the next round, i.e. $anchor_{i+1} = S_{gt} \cup S_{fp}(i)$.
%
%
This strategy can substantially narrow sampling scope, reducing to all positives and only hard negatives, effectively addressing both inter-class and intra-class imbalance. 
Moreover, recent GPUs typically feature a memory up to 12 GB (e.g. NVIDIA Titan Xp), which is scarce when applied to 3D processing with deep CNNs. Training over patches that are much smaller than full-sized images brings a major relief to memory constraint.

In addition, popular deep learning techniques such as stochastic gradient descent and data augmentation (including scaling, flipping and rotating)  are also employed, at the patch level rather than image level. Our loss function is similar to that in the Faster R-CNN work \cite{faster-rcnn}, which is a combination of cross entropy to account for classification loss and smooth L1 for regression loss.

%



%

\newpage

\section{Results and Discussion}

\subsection{Dataset and Experimental Setting}


The LUNA dataset \cite{luna} is a public benchmark widely used for direct comparison between various methods proposed in the literature. 
It contains 888 CT scans, each was annotated by four radiologists, 
where lesions were marked as one of the three classes: 1) non-nodule, 2) nodule $<3mm$, 3) nodule $\geqslant 3mm$. 
Subsequently, each lesion marked as a nodule $\geqslant 3mm$ by at least three radiologists was retained as a ``nodule'' in the ground truth. This results in 1186 nodules in total. 
Then the ones marked by less than three radiologists, or sized $< 3 mm$, or marked as non-nodules, were considered ``irrelevant findings'', and the rest as ``background''.

An experimentation has been carried out on LUNA to evaluate the proposed methodology. A leave-one-out cross-validation scheme was applied, which divided the data into five equal folds; and in each run, a separate network was trained on four folds then tested on the remaining fold. 
In our best experiments, we set patch size $\gamma=96$, IOU thresholds $\eta_{+} = 0.3$, $\eta_{-} = 0.001$, number of online hard negatives $k=5$. The sampling anchors were set to update after every $t=100$ training epochs, and the regression anchors to $A=\left< 3,5,7,10,13,17,22,30,40 \right>$ millimetres ($N_{A}$ denotes the total number of regression anchors). The learning rate was initially set to $lr = 1^{-3}$, and then decayed by a factor of $0.1$ for three times when the loss plateaued. We trained each model 500 epochs in each experiment. A notable point is that all models in this study were trained from scratch, as there was no suitable existing model available for transfer learning.

\subsection{Evaluation with FROC[luna] and  FROC[new]}
On the LUNA standard, detection performance is evaluated via a free receiver operating characteristic (FROC) analysis that
calculates average sensitivity over 7 conditions, respectively [1/8, 1/4, 1/2, 1, 2, 4, 8] FP/scan. 
Table \ref{tab:detection} shows the performance of our methods in comparison to prior work, where our best single-model scored
0.9351, substantially outperforming prior state-of-the-art.
To evaluate the influence of sampling anchors,  an ablation study was conducted where the anchors were fixed to the initial setting without updating:
the same model scored 0.9207 instead. Another ablation without anchor-based patch sampling was also attempted but not successful, as the GPU memory constraint did not allow training on full-sized images. Meanwhile, we tried training models over slightly different hyper-parameters, for example, increasing/reducing several layers, with a slightly different $k$ or $lr$, etc. Many achieved more or less similar performance. An ensemble of several models trained separately achieved a FROC[luna] score of 0.9411. The experiments were not extensive due to time limits and the performance may continue to improve given further hyper-parameter tuning.


\begin{table}[!t]

\caption{Nodule Detection Performance: a comparative view} 
\vspace{-8mm}
\label{tab:detection}

\begin{center}       
\begin{tabular}{|c||c|c|c|c|c|c|c||c||c|} 

\hline
  					&  1/8 & 1/4 & 1/2 & 1 & 2 & 4 & 8 & \textbf{FROC[luna]} & \textbf{FROC[new]} \\
\hline
\hline
Setio et al. \cite{setio}  	& - & - & - & 0.854 & - & 0.901 & - & N/A & N/A \\
\hline
Dou et al. \cite{douqi}  	& 0.659 & 0.745 & 0.819 & 0.865 & 0.906 & 0.933 & 0.946 & 0.8390 & 0.9013 \\
\hline
Ding et al. \cite{ding}   	& 0.748 & 0.853 & 0.887 & 0.922 & 0.938 & 0.944 & 0.946 & 0.8911 & 0.9347 \\
\hline
\hline
Ours (single)  			& 0.815 & 0.868 & 0.952 & 0.968 & 0.981 & 0.981 & 0.981 & 0.9351 & 0.9767 \\
\hline
Ours (ablation)   		& 0.784 & 0.862 & 0.917 & 0.952 & 0.968 & 0.981 & 0.981 & 0.9207 & 0.9670 \\
\hline
Ours (ensemble)   		& 0.827 & 0.871 & 0.958 & 0.976 & 0.982 & 0.982 & 0.987 & 0.9411 & 0.9817 \\

\hline
\end{tabular}
\end{center}
\vspace{-3mm}
\end{table}

Furthermore, we have also started a pilot follow-up study towards clinical practice.
This on-going study involves close discussion with experts from a number of top-tier hospitals in China, regarding empirical considerations in real clinical environments. 
In particular, we identified one of the key issues being the ultimate responsibility of a misdiagnosis, i.e. when a misdiagnosis occurs, 
it would be impractical to hold the machine or the system developer accountable for the patient's loss. 
A more suitable role of the detection system therefore, is not as a substitute to replace radiologists, but a tool to assist them, with improved efficiency.
The detection output is to be reviewed, and revised if necessary, by the radiologist in charge, before a final diagnosis is made.

In this case, we observed that radiologists are generally much more concerned with false negatives, which are difficult to revise, than false positives, which are much easier. In a set of interviews, we noticed that they are generally more tolerant to several FPs in a scan and tend to hold positive opinions towards the trade-off of a small degree of detection precision for an improved sensitivity.  For that reason, we suggest the use of FROC over [1, 2, 4] FP/scan as a better metric.
On the new standard, the methods in comparison generally scored over 0.9, as shown in Table \ref{tab:detection}. Our best single-model scored 0.9767 and the ensemble 0.9817, which not only marks significant improvement over prior art, but also indicates that clinical practice may be close now.



\subsection{Computation Time}
All experimentation was conducted on a standard machine with an Intel Xeon E5-2620 CPU and four NVIDIA Titan Xp GPUs. 
At testing time, the end-to-end workflow on average took around 35s to complete, which is much faster than any two-phase approaches proposed by far.
More specifically, pre-processing took around 20s and network inference took around 15s, which may vary slightly depending on the size of CT image. 
The computational overhead in the training stage on the other hand, is more complex. 
First of all, different models may converge on different numbers of training epochs, and epoches with different sampling anchors may take different periods of time. 
This is because our anchor-based patch sampling introduces more heuristics. 
After each round of training, additional computation will be required for interim evaluation to update the sampling anchors. 
Moreover, given the anchors, the size of patches cropped around them can make significant impact on the training time. 
For instance, training an epoch may take around 5 minutes when $\gamma = 96$, but could increase to 13 minutes when $\gamma = 128$.
In our experiments, it generally took several days to completely train a model from scratch. 
The period can be dramatically shortened if some of the model parameters are re-used when training the next, but not in this study.



\newpage
\section{Conclusion}
In conclusion, we proposed a single-phase single-stage deep learning approach to lung nodule detection in 3D chest CT images, with feature extraction based on a modification of the ResNet and FPN combined, powered by RReLU activation. 
Secondly, we proposed anchor-based patch sampling as a training strategy to address both inter-class and intra-class sample imbalance, improving training efficiency and effectiveness.
Finally, we suggested FROC over [1, 2, 4] FP/scan as the future evaluation criterion, which better suits practical considerations in real clinical environments. Overall, our methods outperformed prior art significantly in terms of both accuracy and speed, scoring FROC[luna] with 0.9351 and FROC[new] with 0.9767 at around 35s per image, using a single model. A simple ensemble on the other hand raised to 0.9411 and 0.9817 respectively.

\end{document}